\documentclass[10pt,twocolumn,letterpaper]{article}

\usepackage{iccv}
\usepackage{times}
\usepackage{epsfig}
\usepackage{graphicx}
\usepackage{amsmath}
\usepackage{amssymb}

\usepackage{multirow,tabularx}
\usepackage{booktabs}
\usepackage{pifont}
\usepackage{floatrow}
\usepackage{xcolor}

\usepackage[breaklinks=true,bookmarks=false]{hyperref}
\newcommand{\cmark}{\ding{51}}%
\newcommand{\xmark}{{\color{red} \ding{55}}}%

\iccvfinalcopy 

\def\iccvPaperID{5810} 

\ificcvfinal\fi

\newcommand*\samethanks[1][\value{footnote}]{\footnotemark[#1]}

\newcommand*\STVQA{\hbox{ST-VQA}}
\begin{document}

\title{Scene Text Visual Question Answering}

\author{Ali Furkan Biten\thanks{Equal contribution.} $^{,1}$ ~ ~ Rub\`{e}n Tito\samethanks{} $^{,1}$ ~ ~ Andres Mafla\samethanks{} $^{,1}$ ~ ~ Lluis Gomez$^{1}$\\ 
Mar\c{c}al Rusi{\~n}ol$^{1}$ ~ ~ Ernest Valveny$^{1}$ ~ ~ C.V. Jawahar$^{2}$ ~ ~ Dimosthenis Karatzas$^{1}$\\
\\
$^1$Computer Vision Center, UAB, Spain ~ ~ $^2$CVIT, IIIT Hyderabad, India \\
{\tt\small \{abiten, rperez, amafla, lgomez, marcal, dimos\}@cvc.uab.es}
}

\maketitle

\ificcvfinal\thispagestyle{empty}\fi

\begin{abstract}
    Current visual question answering datasets do not consider the rich semantic information conveyed by text within an image. In this work, we present a new dataset, \mbox{ST-VQA}, that aims to highlight the importance of exploiting high-level semantic information present in images as textual cues in the Visual Question Answering process.
    We use this dataset to define a series of tasks of increasing difficulty for which reading the scene text in the context provided by the visual information is necessary to reason and generate an appropriate answer.
    We propose a new evaluation metric for these tasks to account both for reasoning errors as well as shortcomings of the text recognition module. In addition we put forward a series of baseline methods, which provide further insight to the newly released dataset, and set the scene for further research.
\end{abstract}

\section{Introduction}

Textual content in man-made environments conveys important high-level semantic information that is explicit and not available in any other form in the scene.
Interpreting written information in man-made environments is essential in order to perform most everyday tasks like making a purchase, using public transportation, finding a place in the city, getting an appointment, or checking whether a store is open or not, to mention just a few.

Text is present in about 50\% of the images in large-scale datasets such as MS Common Objects in Context~\cite{veit2016coco} and the percentage goes up sharply in urban environments. It is thus fundamental to design models that take advantage of these explicit cues.
Ensuring that scene text is properly accounted for is not a marginal research problem, but quite central for holistic scene interpretation models.

The research community on reading systems has made significant advances over the past decade \cite{karatzas2015icdar, gomez2017icdar2017}. The current state of the art in scene text understanding allows endowing computer vision systems with basic reading capacity, although the community has not yet exploited this towards solving higher level problems.


At the same time, current Visual Question Answering (VQA) datasets and models present serious limitations as a result of ignoring scene text content, with disappointing results on questions that require scene text understanding. We therefore consider it is timely to bring together these two research lines in the VQA domain.
\begin{figure}[tp]
\begin{center}
\begin{tabular}{p{0.45\linewidth} p{0.45\linewidth}}
    \includegraphics[width=\linewidth,height=0.75\linewidth]{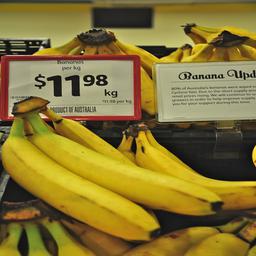}  &
    \includegraphics[width=\linewidth,height=0.75\linewidth]{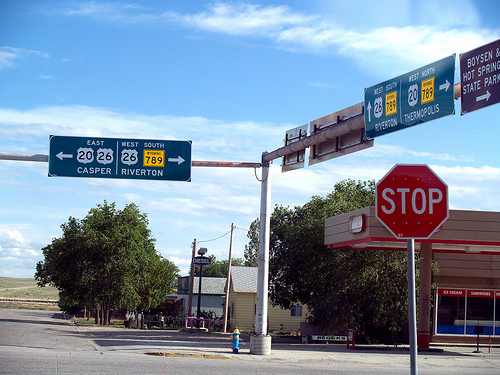}\\
    
    \footnotesize{\fontfamily{qhv}\selectfont \textbf{Q:} What is the price of the bananas per kg?} \par {\color{blue}\footnotesize{\fontfamily{qhv}\selectfont \textbf{A:} \$11.98}}
    &

    \footnotesize{\fontfamily{qhv}\selectfont \textbf{Q:} What does the red sign say?} \par  {\color{blue}\footnotesize{\fontfamily{qhv}\selectfont \textbf{A:} Stop}} \\
    & \\
    \includegraphics[width=\linewidth,height=0.75\linewidth]{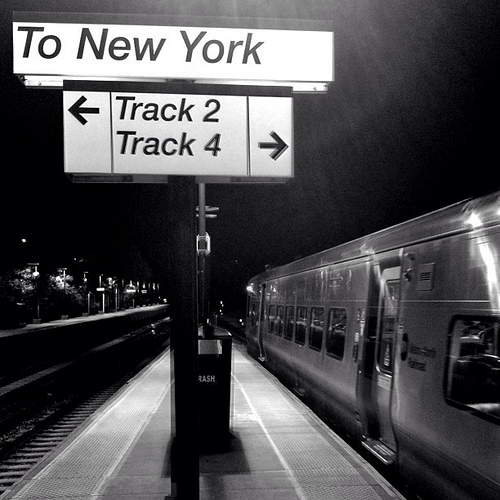}  & \includegraphics[width=\linewidth,height=0.75\linewidth]{2321742.png} \\
    \footnotesize{\fontfamily{qhv}\selectfont \textbf{Q:} Where is this train going?} \par {\color{blue}\footnotesize{\fontfamily{qhv}\selectfont \textbf{A:} To New York \newline \textbf{A:} New York}} &
    \footnotesize{\fontfamily{qhv}\selectfont \textbf{Q:} What is the exit number on the street sign?} \par {\color{blue}\footnotesize{\fontfamily{qhv}\selectfont \textbf{A:} 2 \newline \textbf{A:} Exit 2}}
\end{tabular}
\end{center}
\caption{Recognising and interpreting textual content is essential for scene understanding. In the Scene Text Visual Question Answering (ST-VQA) dataset leveraging textual information in the image is the only way to solve the QA task.}
\label{fig:main}
\vspace{-0.5cm}
\end{figure}
To move towards more human like reasoning, we contemplate that grounding question answering both on the visual and the textual information is necessary.
Integrating the textual modality in existing VQA pipelines is not trivial. On one hand, spotting \textit{relevant} textual information in the scene requires performing complex reasoning about positions, colors, objects and semantics, to localise, recognise and eventually interpret the recognised text in the context of the visual content, or any other contextual information available.
On the other hand, current VQA models work mostly on the principle of classical~\cite{pavlov1960conditioned} and operant (instrumental) conditioning~\cite{skinner1963operant}. Such models, display important dataset biases~\cite{johnson2017clevr} as well as failures in counting~\cite{chattopadhyay2017counting, acharya2019tallyqa}, comparing and identifying attributes. These limitations make current models unsuitable to directly integrate scene text information which is often orthogonal and uncorrelated to the visual statistics of the image.

To this end, in this work we propose a new dataset, called \textit{Scene Text Visual Question Answering} (ST-VQA) where the questions and answers are attained in a way that questions can only be answered based on the text present in the image. We consciously draw the majority (85.5\%) of ST-VQA images from datasets that have generic question/answer pairs that can be combined with ST-VQA to establish a more generic, holistic VQA task. 
Some sample images and questions from the collected dataset are shown in \autoref{fig:main}.

Additionally, we introduce three tasks of increasing difficulty that simulate different degrees of availability of contextual information. Finally, we define a new evaluation metric to better discern the models' answering ability, that employs the Levenshtein distance~\cite{levenshtein1966binary} to account both for reasoning errors as well as shortcomings of the text recognition subsystem \cite{gomez2017icdar2017}. The dataset, as well as performance evaluation scripts and an online evaluation service are available through the ST-VQA Web portal\footnote{\url{https://rrc.cvc.uab.es/?ch=11}}.


\section{Related Work}
The task of text detection and recognition in natural images sets the starting point for a generalized VQA system that can integrate textual cues towards complete scene understanding. The most common approach in the reading systems community consists of two steps, text detection and recognition. Several works have been proposed addressing text detection such as ~\cite{liao2017textboxes, liao2018textboxes++, zhou2017east, he2017deep} which are mostly based on Fully Convolutional Neural Networks. 

Text recognition methods such as the one presented in~\cite{jaderberg2016reading} propose recognizing text at the word level as a classification problem (word spotting) from a $90K$ English words vocabulary. 
Approaches that use Connectionist Temporal Classification have also been widely used in scene text recognition, in works such as~\cite{shi2016robust, borisyuk2018rosetta, yin2015multi, gao2017reading, liu2018fots}, among others. Later works focus towards end-to-end architectures such as the ones presented by~\cite{busta2017deep, lyu2018mask, he2018end}, which mostly consist of an initial Convolutional Neural Network (CNN) that acts as an encoder and a Long Short Term Memory (LSTM) combined with attention that acts as the decoder.

Visual Question Answering (VQA) aims to come up with an answer to a given natural language question about the image. Since its introduction, VQA has received a lot of attention from the Computer Vision community~\cite{antol2015vqa, gao2015you, ren2015exploring, goyal2017making, johnson2017clevr, agrawal2018don} facilitated by access to large-scale datasets that allow the training of VQA models~\cite{antol2015vqa, goyal2017making, krishna2017visual, yu2015visual, tapaswi2016movieqa, malinowski2014multi}. Despite VQA's popularity, none of the existing datasets except TextVQA (reviewed separately next) consider textual content, while in our work, exploiting textual information found in the images is the only way to solve the VQA task. 

Related to the task proposed in this paper, are the recent works of Kafle et~al.~\cite{kafle2018dvqa} and Kahou et~al.~\cite{kahou2017figureqa} on question answering for bar charts and diagrams, the work of Kise at al.~\cite{kise2005document} on QA for machine printed document images, and the work of Kembhavi et~al.~\cite{kembhavi2017you} on textbook question answering. The Textbook Question Answering (TQA) dataset~\cite{kembhavi2017you} aims at answering multimodal questions given a context of text, diagrams and images, but textual information is provided in computer readable format. This is not the case for the diagrams and charts of the datasets proposed in~\cite{kafle2018dvqa,kahou2017figureqa}, meaning that models require some sort of text recognition to solve such QA tasks. However, the text found on these datasets is rendered in standard font types and with good quality, and thus represents a less challenging setup than the scene text used in our work. 

TextVQA~\cite{singh2019} is a concurrent work to the one presented here. Similarly to ST-VQA, TextVQA proposes an alternative dataset for VQA which  requires reading and reasoning about scene text. Additionally,~\cite{singh2019} also introduces a novel architecture that combines a standard VQA model~\cite{singh2018pythia} and an independently trained OCR module~\cite{borisyuk2018rosetta} with a ``copy'' mechanism, inspired by pointer networks~\cite{vinyals2015pointer,gulcehre2016pointing}, which allows to use OCR recognized words as predicted answers if needed. Both TextVQA and ST-VQA datasets are conceptually similar, although there are important differences in the implementation and design choices. We offer here a high-level summary of key differences, while section~\ref{sec:AnalysisAndComparison} gives a quantitative comparison between the two datasets.

In the case of ST-VQA, a number of different source image datasets were used, including scene text understanding ones, while in the case of TextVQA all images come from a single source, the Open Images dataset.
To select the images to annotate for the ST-VQA, we explicitly required a minimum amount of two text instances to be present, while in TextVQA images were sampled on a category basis, emphasizing categories 
that are expected to contain text.
In terms of the questions provided, ST-VQA focuses on questions that can be answered unambiguously directly using part of the image text as answer, while in TextVQA any question requiring reading the image text is allowed. 




Despite the differences, the two datasets are highly complementary as the image sources used do not intersect with each other, creating an opportunity for transfer learning between the two datasets and maybe combining data for training models with greater generalization capabilities.


\section{ST-VQA Dataset}
\subsection{Data Collection}
In this section we describe the process for collecting images, questions and answers for the ST-VQA dataset, and offer an in-depth analysis of the collected data. Subsequently, we detail the proposed tasks and introduce the evaluation metric.

\textbf{Images:} The ST-VQA dataset comprises $23,038$ images sourced from a combination of public datasets that include both scene text understanding datasets as well as generic computer vision ones.
In total, we used six different datasets, namely: ICDAR 2013\cite{karatzas2013icdar} and ICDAR2015\cite{karatzas2015icdar}, ImageNet \cite{deng2009imagenet}, VizWiz\cite{gurari2018vizwiz}, IIIT Scene Text Retrieval\cite{MishraICCV13_IIIT_text}, Visual Genome \cite{krishna2017visual} and COCO-Text \cite{veit2016coco}. A key benefit of combining images from various datasets is the reduction of dataset bias such as selection, capture and negative set bias which have been shown to exist in popular image datasets\cite{khosla2012undoing}. Consequently, the combination of datasets results in a greater variability of questions. 
To automatically select images to define questions and answers, we use an end-to-end single shot text retrieval architecture~\cite{gomez2018single}. We automatically select all images that contain at least 2 text instances thus ensuring that the proposed questions contain at least 2 possible options as an answer. The final number of images and questions per dataset can be found in \autoref{tab:dbstats1}.

\begin{table}[ht]
\begin{tabularx}{0.9\linewidth}{X r r}
    \toprule
    Original Dataset & Images & Questions \\ 
    \midrule
    Coco-text & 7,520 & 10,854 \\ 
    Visual Genome & 8,490 & 11,195 \\
    VizWiz  & 835  & 1,303 \\ 
    ICDAR   & 1,088 & 1,423 \\ 
    ImageNet & 3,680 & 5,165 \\ 
    IIIT-STR & 1,425 & 1,890 \\
    \midrule
    Total & 23,038 & 31,791 \\ 
    \bottomrule
\end{tabularx}
\caption{Number of images and questions gathered per dataset.}
\label{tab:dbstats1}
\end{table}

\textbf{Question and Answers:} The ST-VQA dataset comprises $31,791$ questions. To gather the questions and answers of our dataset, we used the crowd-sourcing platform Amazon Mechanical Turk (AMT). During the collection of questions and answers, we encouraged workers to come up with closed-ended questions that can be unambiguously answered with text found in the image, prohibiting them to ask yes/no questions or questions that can be answered only based on the visual information. 

The process of collecting question and answer pairs consisted of two steps. First, the workers were given an image along with instructions asking them to come up with a question that can be answered using the text found in the image. The workers were asked to write up to three question and answer pairs. Then, as a verification step, we perform a second AMT task that consisted of providing different workers with the image and asking them to respond to the previously defined question.
We filtered the questions for which we did not obtain the same answer in both steps, in order to remove ambiguous questions. The ambiguous questions were checked by the authors and corrected if necessary, before being added to the dataset. In some cases both answers were deemed correct and accepted, therefore ST-VQA questions have up to two different valid answers.

In total, the proposed ST-VQA dataset comprises $23,038$ images with $31,791$ questions/answers pair separated into $19,027$ images - $26,308$ questions for training and $2,993$ images - $4,163$ questions for testing. We present examples of question and answers of our dataset in \autoref{fig:main}.

\subsection{Analysis and Comparison with TextVQA}
\label{sec:AnalysisAndComparison}

\begin{figure}[h]
    \centering
    \includegraphics[width=\linewidth]{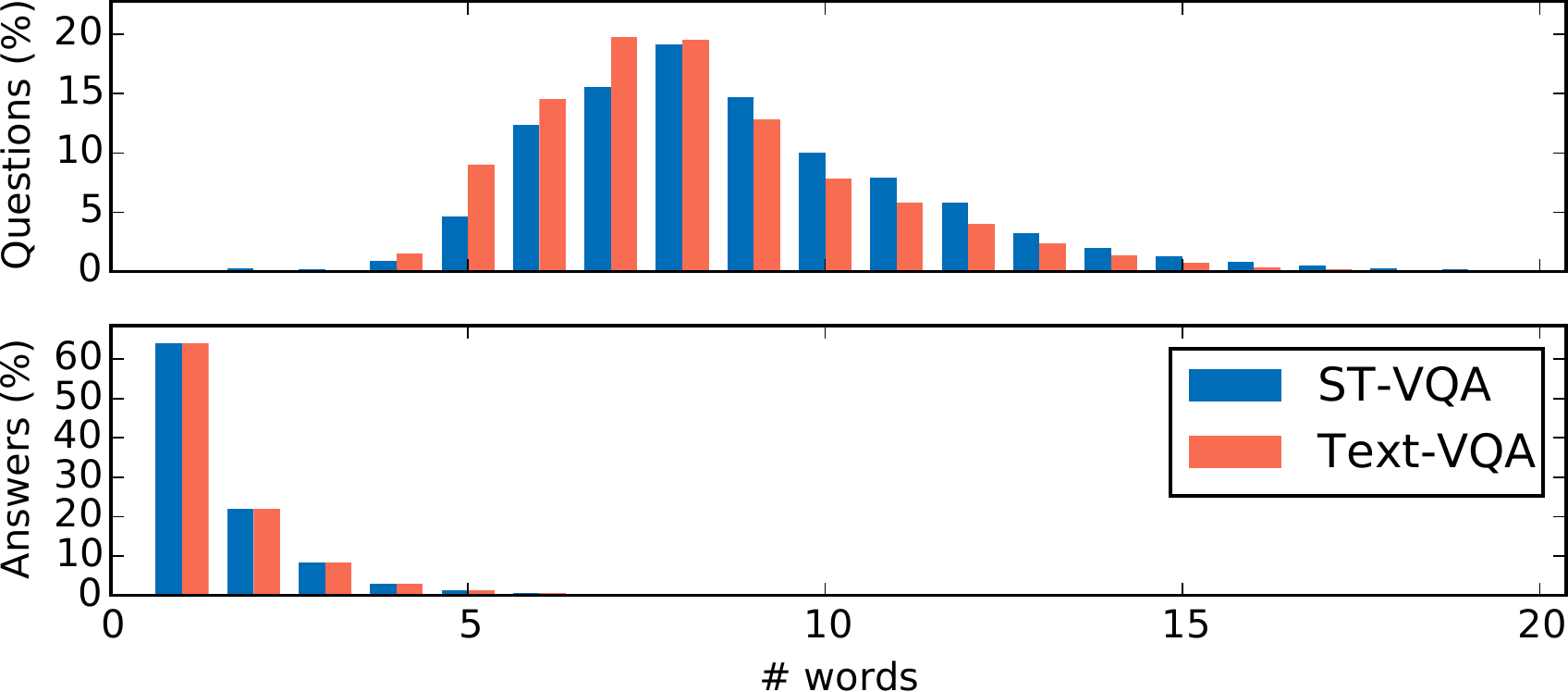}
    \caption{Percentage of questions (top) and answers (bottom) that contain a specific number of words.}
    \label{fig:question_stats}
    \label{fig:answer_stats}
\end{figure}

In \autoref{fig:question_stats} we provide the length distribution for the gathered questions and answers of the ST-VQA datasets, in comparison to the recently presented TextVQA. It can be observed that the length statistics of the two datasets are closely related.

\begin{figure}[h]
    \centering
    \includegraphics[width=\linewidth]{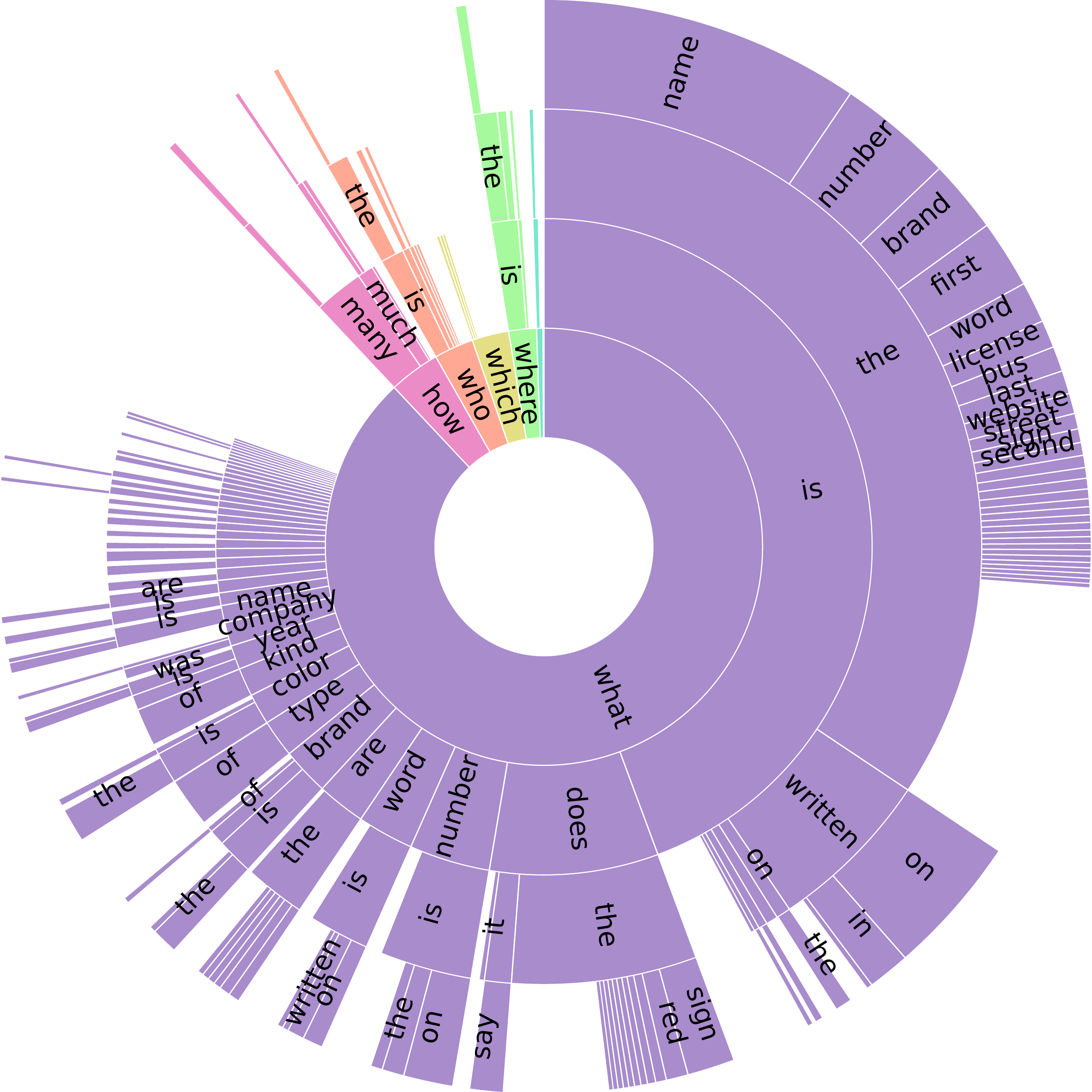}
    \caption{Distribution of questions in the ST-VQA train set by their starting 4-grams (ordered from center to outwards). Words with a small contribution are not shown for better visualization.}
    \label{fig:question_tokens}
\end{figure}

To further explore the statistics of our dataset, \autoref{fig:question_tokens} visualises how the ST-VQA questions are formed.
As it can be appreciated, our questions start with ``What, Where, Which, How and Who''. A considerable percentage starts with ``What'' questions, as expected given the nature of the task. A critical point to realize however, is that the questions are not explicitly asking for specific text that appears in the scene; rather they are formulated in a way that requires to have certain prior world knowledge/experience. For example, some of the \textit{``what''} questions inquire about a brand, website, name, bus number, etc., which require some explicit knowledge about what a brand or website is. 

\begin{figure*}[t]
    \centering
    \includegraphics[width=\textwidth]{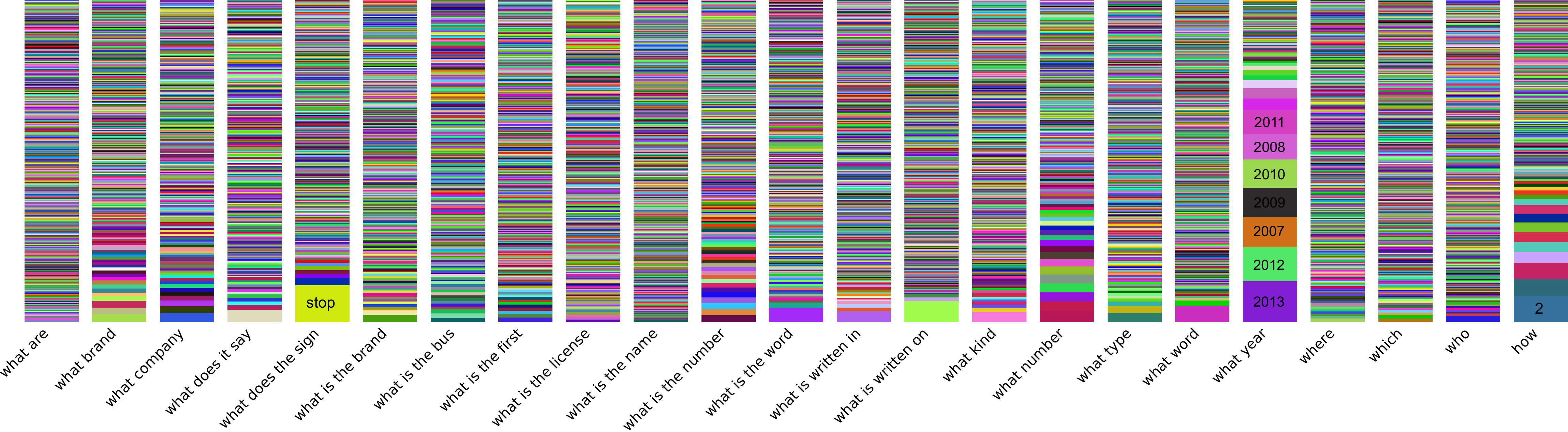}
    \caption{Distribution of answers for different types of questions in the ST-VQA train set. Each color represents a different unique answer.}
    \label{fig:answers_tokens}
\end{figure*}

There has been a lot of effort to deal with the language prior inside the datasets~\cite{goyal2017making, johnson2017clevr, zhang2016yin}. One of the reasons for having language priors in datasets is the uneven distribution of answers in the dataset. In VQA v1~\cite{antol2015vqa}, since the dataset is formed from the images of MSCOCO~\cite{lin2014microsoft}, the answers to the question of ``what sport ...'' are \textit{tennis} and \textit{baseball} over 50\%. Another example is the question ``is there ...'', having \textit{yes} as an answer in over $70\%$ of the cases. As can be seen from \autoref{fig:answers_tokens}, our dataset apart from the ``sign'' and ``year'' questions follows a uniform distribution for the answers, reducing the risk of language priors while having a big vocabulary for the answers.




To put ST-VQA in perspective, VQA 2.0~\cite{goyal2017making}, the biggest dataset in the community, contains 1.1 million questions out of which only 8k, corresponding to less than $1\%$ of the total questions, requires reading the text in the image. The TextVQA~\cite{singh2019} dataset on the other hand comprises $28,408$ images paired with $45,336$ questions. 

As a result of the different collection procedures followed, all ST-VQA questions can be answered unambiguously directly using the text in the image, while in the case of TextVQA reportedly $39\%$ ($18k$) of the answers do not contain any of the OCR tokens~\footnote{Presentation of the TextVQA Challenge, CVPR 2019}. This might be either due to the type of the questions defined, or due to shortcomings of the employed text recognition engine.


The fact that ST-VQA answers are explicitly grounded to the scene text, allows us to collect a single answer per question
To consider an answer as correct, we introduce a soft metric that requires it to have a small edit distance to the correct answer (see section~\ref{sec:Evaluation}), factoring this way in the evaluation procedure the performance of the text recognition sub-system. In the case of TextVQA, $10$ answers are collected per question and any answer supported by at least three subjects is considered correct.
In order to better understand the effects of our approach compared to collecting multiple responses like in TextVQA, we performed an experiment collecting $10$ answers for a random subset of $1,000$ \STVQA~questions. Our analysis showed that in $84.1\%$ of the cases there is agreement between the majority of subjects and the original answer. The same metric for TextVQA is $80.3\%$, confirming that defining a single unambiguous answer results in similarly low ambiguity at evaluation time.

\subsection{Tasks}
\label{subsec:tasks}
We define 3 novel tasks, suitable for the ST-VQA dataset, namely ``strongly contextualised'', ``weakly contextualised'' and ``open vocabulary''.

The proposed differentiation of tasks can be interpreted by how humans make use of prior knowledge to argue about their current situation. Such prior knowledge in \STVQA~is provided as a dictionary, different for each task. Similar approaches using dynamic per-image dictionaries have been used for DVQA in~\cite{kafle2018dvqa} and for scene text understanding in~\cite{karatzas2015icdar}.
Our formulation of the tasks is inspired by the previous notions and the difficulty per task increases gradually. In the strongly contextualised task we capture this prior knowledge by creating a dictionary \textit{per image} for the specific scenario depicted. In the weakly contextualised task we provide a \textit{single dictionary} comprising all the words in the answers of the dataset. Finally, for the open dictionary task, we treat the problem as tabula rasa where no a priori and no external information is available to the model.

For the strongly contextualised task (1), following the standard practice used for end-to-end word spotting~\cite{karatzas2013icdar, karatzas2015icdar, wang2010word}, we create a dictionary per image that contains the words that appear in the answers defined for questions on that image, along with a series of distractors. The distractors are generated in two ways. On one hand, they comprise instances of scene text as returned by a text recogniser applied on the image. On the other hand, they comprise words obtained by exploiting the semantic understanding of the scene, in the form of the output of a dynamic lexicon generation model \cite{patel2016dynamic,gomez2017self}.
The dictionary for the strongly contextualised task is $100$ words long and defined per image. 

In the weakly contextualised task (2), we provide a unique dictionary of $30,000$ words for all the datasets' images which is formed by collecting all the $22k$ ground truth words plus $8k$ distractors generated in the same way as in the previous task. 
Finally for the open dictionary task (3), we provide no extra information thus we can consider it as an open-lexicon task. 

By proposing the aforementioned tasks the VQA problem is conceived in a novel manner that has certain advantages. First, it paves the way for research on automatically processing and generating such prior information, and its effect on the model design and performance. 
Second, it provides an interesting training ground for end-to-end reading systems, where the provided dictionaries can be used to prime text spotting methods.

\subsection{Evaluation and Open Challenge}
\label{sec:Evaluation}




Since the answers of our dataset are contained within the text found in the image, which is dependent on the accuracy of the OCR being employed, the classical evaluation metric of VQA tasks is not optimum for our dataset, e.g. if the model reasons properly about the answer but makes a mistake of a few characters in the recognition stage, like in~\autoref{fig:main2} (first row, third column), the typical accuracy score would be $0$. However, the metric we propose named Average Normalized Levenshtein Similarity (ANLS) would give an intermediate score between 0.5 and 1 that will softly penalise the OCR mistakes. 
Thus, a motivation of defining a metric that captures OCR accuracy as well as model reasoning is evident. To this end, in all 3 tasks we use the normalized Levenshtein similarity~\cite{levenshtein1966binary} as an evaluation metric. More formally, we define \textrm{ANLS} as follows:

\begin{equation}
\label{eq:ANLS}
\textrm{ANLS} = \frac{1}{N} \sum_{i=0}^{N} \left(\max_{j}  s(a_{ij}, o_{q_i}) \right) \\
\end{equation}
\vspace{-0.2cm}
\[
  s(a_{ij}, o_{q_i}) =
  \begin{cases}
            (1 - NL(a_{ij}, o_{q_i})) & \text{if \textrm{ $NL(a_{ij},o_{q_i}) < \tau$}} \\
                                    $0$ & \text{if \textrm{ $NL(a_{ij},o_{q_i})$  $\geqslant$  $\tau$}} 
  \end{cases}
\]

\noindent
where $N$ is the total number of questions in the dataset, $M$ is the total number of GT answers per question, $a_{ij}$ are the ground truth answers where $i = \{0, ..., N\}$, and $j = \{0, ..., M\}$, and $o_{q_i}$ is the network's answer for the $i^{th}$ question $q_i$.   $NL(a_{ij}, o_{q_i})$ is the normalized Levenshtein distance between the strings $a_{ij}$ and $o_{q_i}$ (notice that the normalized Levenshtein distance is a value between $0$ and $1$). We define a threshold $\tau = 0.5$ that penalizes metrics larger than this value, thus the final score will be $0$ if the $NL$ is larger than $\tau$. The intuition behind the threshold is that if an output has an edit distance of more than $0.5$ to an answer, meaning getting half of the answer wrong, we reason that the output is the wrong text selected from the options as an answer. Otherwise, the metric has a smooth response that can gracefully capture errors in text recognition.

In addition, we provide an online service
where the open challenge was hosted~\cite{biten2019icdar}, that researchers can use to evaluate their methods against a public validation/test dataset. 

\section{Baselines and Results}
The following section describes the baselines employed in this work as well as an analysis of the results obtained in the experiments conducted. 
The proposed baselines help us to showcase the difficulty of the proposed dataset and its tasks. Aside from baselines designed to exploit all the information available (visual information, scene text, and the question), we have purposely included baselines that ignore one or more of the available pieces of information in order to establish lower bounds of performance. The following baselines are employed to evaluate the datasets: \par
\textbf{Random:} 
As a way of assessing aimless chance, we return a random word from the dictionary provided for each task (see section \ref{subsec:tasks} for more detail).

\textbf{Scene Text Retrieval:}
This baseline leverages a single shot CNN architecture \cite{gomez2018single} that predicts at the same time bounding boxes and a Pyramidal Histogram Of Characters (PHOC)~\cite{almazan2014word}. The PHOC is a compact representation of a word that considers the spatial location of each character to construct the resulting encoding. This baseline ignores the question and any other visual information of the image.

We have defined two approaches: the first (``STR retrieval'') uses the specific task dictionaries as queries to a given image, and the top-1 retrieved word is returned as the answer; the second one (``STR bbox''), follows the intuition that humans tend to formulate questions about the largest text instance in the image. We take the text representation from the biggest bounding box found and then find the nearest neighbor word in the corresponding dictionaries.

\textbf{Scene Image OCR:}
A state of the art text recognition model~\cite{he2018end} is used to process the test set images. The detected text is ranked according to the confidence score and the closest match between the most confident text detection and the provided vocabularies for task 1 and task 2 is used as the answer. In task 3 the most confident text detection is adopted as the answer directly.

\begin{table*}[h]
\begin{tabularx}{\linewidth}{X | c c c | c c c c c c c c}
\toprule
 &   &  &  & \multicolumn{2}{c}{Task 1} & \multicolumn{2}{c}{Task 2} & \multicolumn{2}{c}{Task 3} & \multicolumn{2}{c}{Upper bound}\\ 
Method with &  OCR & Q & V & ANLS & Acc.  & ANLS & Acc.  & ANLS & Acc. & ANLS & Acc.\\
\midrule
Random                                      & \xmark & \xmark & \xmark & 0.015 & 0.96 & 0.001 & 0.00 & 0.00 & 0.00 & - & -\\
STR~\cite{gomez2018single} (retrieval) & \cmark & \xmark & \xmark & \textbf{0.171} & \textbf{13.78} & 0.073 & 5.55 & - & - & 0.782  & 68.84\\
STR~\cite{gomez2018single} (bbox) & \cmark & \xmark & \xmark & 0.130 & 7.32 & 0.118 & 6.89 & 0.128 & 7.21 & -  & -\\
Scene Image OCR~\cite{he2018end}                  & \cmark & \xmark & \xmark & 0.145 & 8.89 & 0.132 & 8.69 & \textbf{0.140} & 8.60 & -  & -\\

SAAA~\cite{kazemi2017show} (1k cls)          & \xmark & \cmark & \cmark & 0.085 & 6.36 & 0.085 & 6.36 & 0.085 & 6.36 & 0.571 & 31.96\\
SAAA+STR (1k cls)                            & \cmark & \cmark & \cmark & 0.091 & 6.66 & 0.091 & 6.66 & 0.091 & 6.66 & 0.571 & 31.96\\
SAAA~\cite{kazemi2017show} (5k cls)          & \xmark & \cmark & \cmark & 0.087 & 6.66 & 0.087 & 6.66 & 0.087 & 6.66 & 0.740 & 41.03\\ 
SAAA+STR (5k cls)                            & \cmark & \cmark & \cmark & 0.096 & 7.41 & 0.096 & 7.41 & 0.096 & 7.41 &  0.740 & 41.03\\
SAAA~\cite{kazemi2017show} (19k cls)         & \xmark & \cmark & \cmark & 0.084 & 6.13 & 0.084 & 6.13 & 0.084 & 6.13 & 0.862 & 52.31\\
SAAA+STR (19k cls)                           & \cmark & \cmark & \cmark & 0.087 & 6.36 & 0.087 & 6.36 & 0.087 & 6.36 & 0.862 & 52.31\\
QA+STR (19k cls)                            & \cmark & \cmark & \xmark & 0.069 & 4.65 & 0.069 & 4.65 & 0.069 & 4.65 & 0.862 & 52.31\\

SAN(LSTM)~\cite{yang2016stacked} (5k cls)   & \xmark & \cmark & \cmark & 0.102 & 7.78 & 0.102 & 7.78 & 0.102 & 7.78 & 0.740 & 41.03\\
SAN(LSTM)+STR (5k cls)                      & \cmark & \cmark & \cmark & 0.136 & 10.34 & \textbf{0.136} & \textbf{10.34} & 0.136 & \textbf{10.34} & 0.740 & 41.03\\
SAN(CNN)+STR (5k cls)                       & \cmark & \cmark & \cmark & 0.135 & 10.46 & \textbf{0.135} & \textbf{10.46} & 0.135 & \textbf{10.46} & 0.740 & 41.03\\
\bottomrule
\end{tabularx}
\caption{Baseline results comparison on the three tasks of ST-VQA dataset. We provide Average Normalized Levenshtein similarity (ANLS) and Accuracy for different methods that leverage OCR, Question (Q) and Visual (V) information.}
\label{tab:baseline_results}
\end{table*}

\textbf{Standard VQA models:}
We evaluate two standard VQA models. The first one, named ``Show, Ask, Attend and Answer''~\cite{kazemi2017show} (SAAA), 
consists of a CNN-LSTM architecture. On one hand, a ResNet-152~\cite{he2016deep} is used to extract image features with dimension $14\times14\times2048$, while the question is tokenized and embedded by using a multi-layer LSTM. On top of the combination of image features and the question embedding, multiple attention maps (glimpses) are obtained. The result of the attention glimpses over the image features and the last state of the LSTM is concatenated and fed into two fully connected layers to obtain the distribution of answer probabilities according to the classes. We optimize the model with the Adam optimizer \cite{kingma2014adam} with a batch size of $128$ for $30$ epochs. The starting learning rate is $0.001$ which decays by half every 50\textit{K} iterations.

The second model, named ``Stacked Attention Networks''~\cite{yang2016stacked} (SAN), uses a pre-trained VGGN~\cite{simonyan2014very} CNN to obtain image features with shape $14\times14\times512$. Two question encoding methods are proposed, one that uses an LSTM and another that uses a CNN, both of them yielding similar results according to the evaluated dataset. The encoded question either by a CNN or LSTM is used along with the image features to compute two attention maps, which later are used with the image features to output a classification vector. We optimize the model with a batch size of $100$ for $150$ epochs. The optimizer used is RMSProp with a starting learning rate of $0.0003$ and a decay value of $0.9999$.

Overall, three different experiments are proposed according to the output classification vector. The first, is formed by selecting the most common 1k answer strings in the ST-VQA training set as in~\cite{antol2015vqa}. For the second one, we selected the 5k most common answers so that we can see the effect of a gradual increase of the output vector in the two VQA models. In the third one, all the answers found in the training set are used ($19,296$) to replicate the wide range vocabulary of scene-text images and to capture all the answers found in the training set.

\textbf{Fusing Modalities - Standard VQA Models + Scene Text Retrieval:}
Using the previously described VQA models, the purpose of this baseline is to combine textual features obtained from a scene text retrieval model with existing VQA pipelines. To achieve this, we use the model from \cite{gomez2018single} and we employ the output tensor before the non-maximal suppression step (NMS) is performed. The most confident PHOC predictions above a threshold are selected relative to a single grid cell. The selected features form a tensor of size $14\times14\times609$, which is concatenated with the image features before the attention maps are calculated on both previously described VQA baselines. Afterwards the attended features are used to output a probability distribution over the classification vector. The models are optimized using the same strategy described before.

\begin{figure*}[h]
    \centering
    \includegraphics[width=\linewidth]{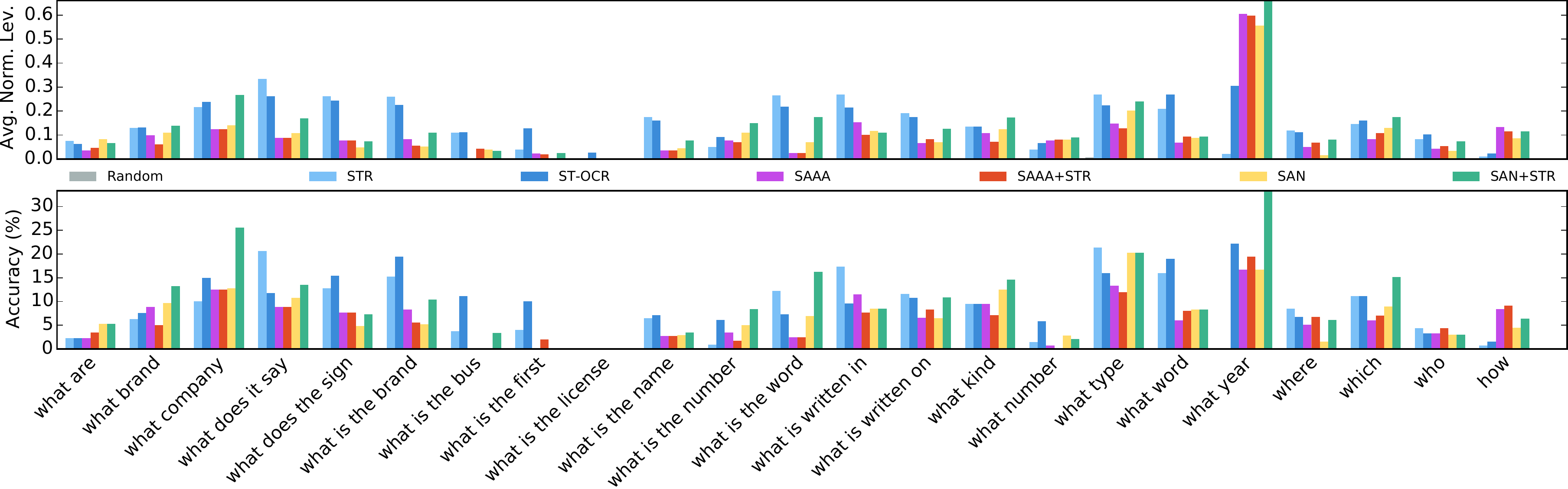}
    \caption{Results of baseline methods in the open vocabulary task of ST-VQA by question type.}
    \label{fig:results_by_type}
\end{figure*}

\subsection{Results}

The results of all provided baselines according to the defined tasks are summarized in~\autoref{tab:baseline_results}. As a way to compare the proposed Average Normalized Levenshtein Similarity (ANLS) metric, we also calculate the accuracy for each baseline. The accuracy is calculated by counting the exact matches between the model predictions and collected answers as is the standard practice in the VQA literature. 

The last column in~\autoref{tab:baseline_results}, upper bound, shows the maximum possible score that can be achieved depending on the method evaluated. 
The upper bound accuracy for standard VQA models is the percentage of questions where the correct answer is part of the models’ output vocabulary, while the upper bound ANLS is calculated by taking as answer the closest word (output class) in terms of Levenshtein distance to the correct answer. 
In the case of the Scene Text Retrieval (STR retrieval) \cite{gomez2018single} model the upper bound is calculated by assuming that the correct answer is a single word and that this word is retrieved by the model as the top-1 among all the words in the provided vocabularies. 



In~\autoref{tab:baseline_results} we appreciate that standard VQA models that disregard textual information from the image achieve similar scores, ranging between 0.085 to 0.102 ANLS, or 6.36\% to 7.78\% accuracy. One relevant point is that although in VQA v1~\cite{antol2015vqa} the SAAA~\cite{kazemi2017show} model is known to outperform  SAN~\cite{yang2016stacked}, in our dataset the effect found is the opposite, due to the fact that our dataset and task outline is different in its nature compared to VQA v1. 

Another important point is that the SAAA model increases both its accuracy and ANLS score when using a larger classification vector size, from 1k to 5k classes; however, going from 5k to 19k classes the results are worse, suggesting that learning such a big vocabulary in a classification manner is not feasible.  

\begin{figure*}[t]
\begin{center}
\begin{tabular}{p{0.23\linewidth} p{0.23\linewidth} p{0.23\linewidth} p{0.23\linewidth}}
    \includegraphics[width=\linewidth,height=0.75\linewidth]{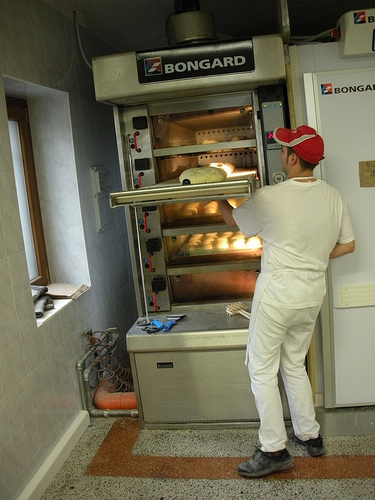}  &
    \includegraphics[width=\linewidth,height=0.75\linewidth]{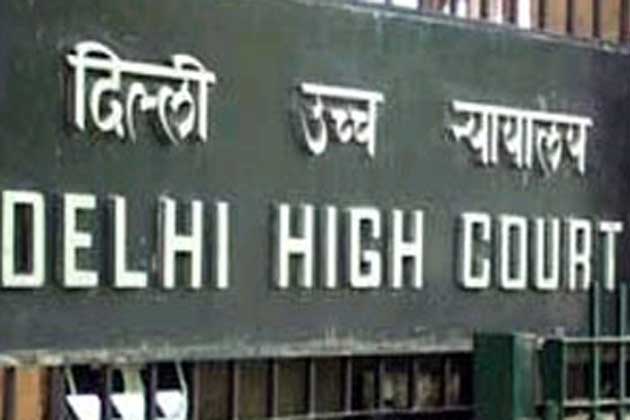} & 
    \includegraphics[width=\linewidth,height=0.75\linewidth]{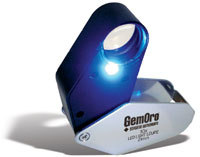}  &
    \includegraphics[width=\linewidth,height=0.75\linewidth]{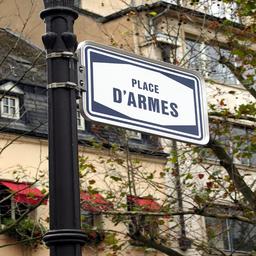} \\

    \footnotesize{\fontfamily{qhv}\selectfont \textbf{Q:} What brand are the machines?} \par {\color{blue}\footnotesize{\fontfamily{qhv}\selectfont \textbf{A:} bongard}} \par
    {\color{red}\footnotesize{\fontfamily{qhv}\selectfont \textbf{SAN(CNN)+STR:} ray}} \par
    {\color{red}\footnotesize{\fontfamily{qhv}\selectfont \textbf{SAAA+STR:} ray}} \par
    {\color{orange}\footnotesize{\fontfamily{qhv}\selectfont \textbf{Scene Image OCR:} zbongard}} \par
    {\color{red}\footnotesize{\fontfamily{qhv}\selectfont \textbf{STR (bbox):} 1}} &
    
    \footnotesize{\fontfamily{qhv}\selectfont \textbf{Q:} Where is the high court located?}  \par {\color{blue}\footnotesize{\fontfamily{qhv}\selectfont \textbf{A:} delhi}} \par
    {\color{green}\footnotesize{\fontfamily{qhv}\selectfont \textbf{SAN(CNN)+STR:} delhi}}\par
    {\color{green}\footnotesize{\fontfamily{qhv}\selectfont \textbf{SAAA+STR:} delhi}} \par
    {\color{red}\footnotesize{\fontfamily{qhv}\selectfont \textbf{Scene Image OCR:} high}} \par
    {\color{green}\footnotesize{\fontfamily{qhv}\selectfont \textbf{STR (bbox):} delhi}} &

    \footnotesize{\fontfamily{qhv}\selectfont \textbf{Q:} What does the black label say?}   \par {\color{blue}\footnotesize{\fontfamily{qhv}\selectfont \textbf{A:} GemOro}} \par
    {\color{red}\footnotesize{\fontfamily{qhv}\selectfont \textbf{SAN(CNN)+STR:} st. george ct.}} \par
    {\color{red}\footnotesize{\fontfamily{qhv}\selectfont \textbf{SAAA+STR:} esplanade}} \par
    {\color{orange}\footnotesize{\fontfamily{qhv}\selectfont \textbf{Scene Image OCR:} gemors}} \par
    {\color{red}\footnotesize{\fontfamily{qhv}\selectfont \textbf{STR (bbox):} genoa}} &

    \footnotesize{\fontfamily{qhv}\selectfont \textbf{Q:} What's the street name?}  \par {\color{blue}\footnotesize{\fontfamily{qhv}\selectfont \textbf{A:} place d'armes}} \par
    {\color{red}\footnotesize{\fontfamily{qhv}\selectfont \textbf{SAN(CNN)+STR:} 10th st}} \par
    {\color{red}\footnotesize{\fontfamily{qhv}\selectfont \textbf{SAAA+STR:} ramistrasse}} \par
    {\color{orange}\footnotesize{\fontfamily{qhv}\selectfont \textbf{Scene Image OCR:} d'armes}} \par
    {\color{red}\footnotesize{\fontfamily{qhv}\selectfont \textbf{STR (bbox):} dames}} \\
     
    \includegraphics[width=\linewidth,height=0.75\linewidth]{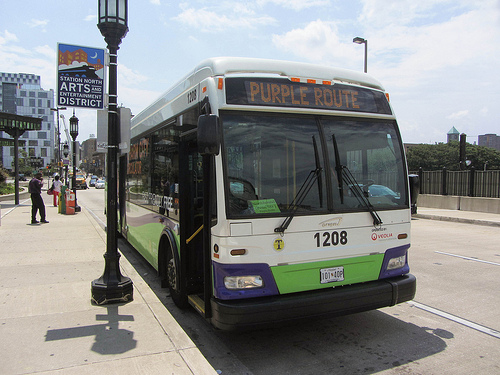}  &
     \includegraphics[width=\linewidth,height=0.75\linewidth]{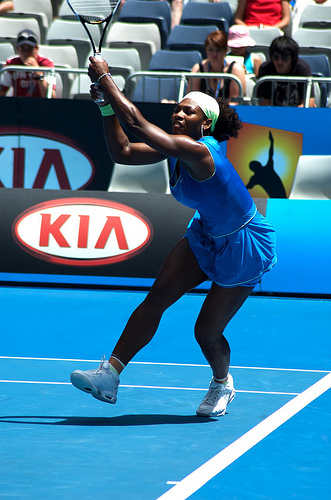} &
     \includegraphics[width=\linewidth,height=0.75\linewidth]{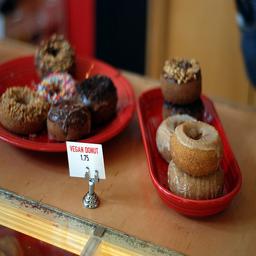} &
     \includegraphics[width=\linewidth,height=0.75\linewidth]{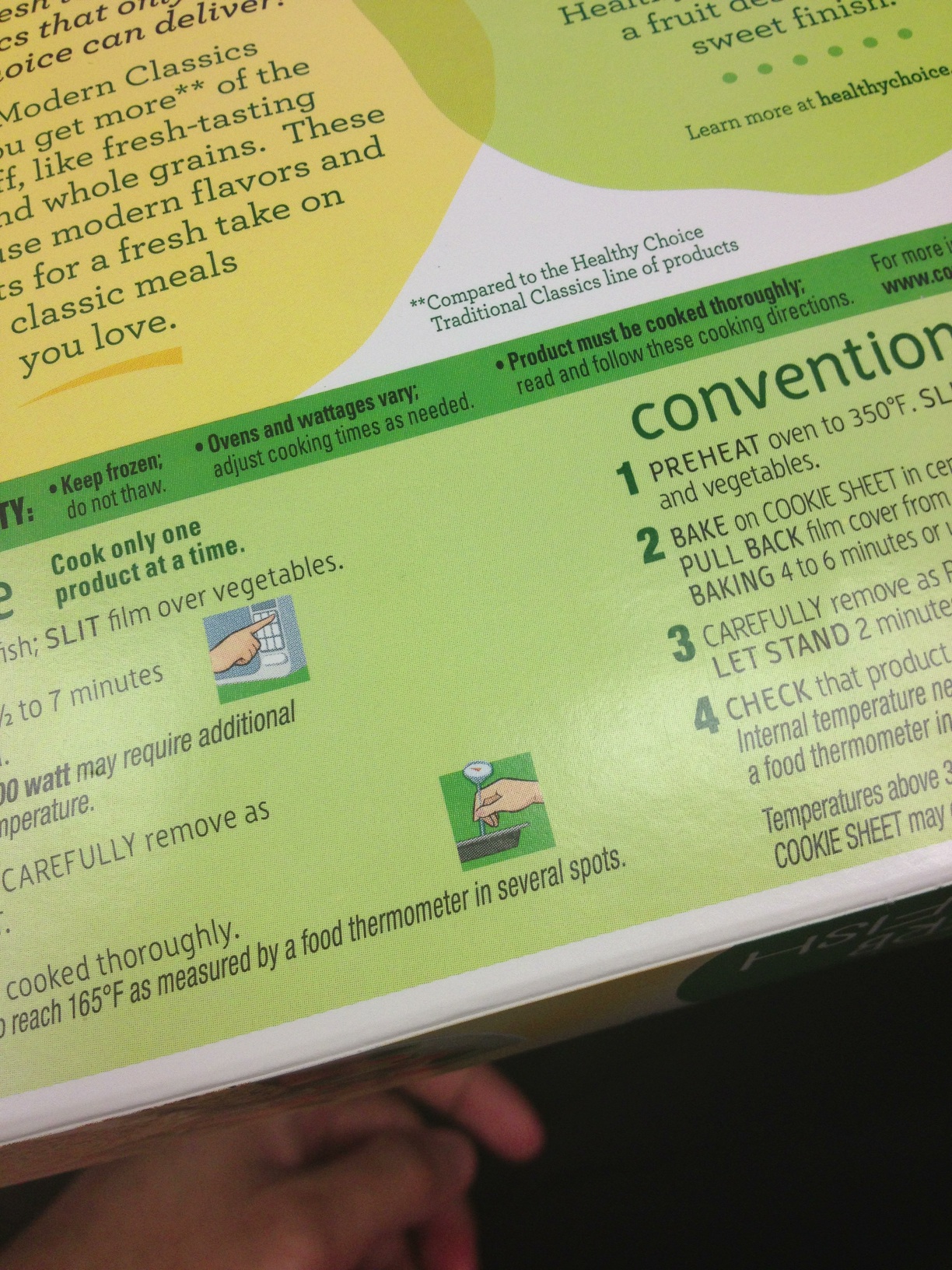} \\

    \footnotesize{\fontfamily{qhv}\selectfont \textbf{Q:} What is the route of the bus?}   \par {\color{blue}\footnotesize{\fontfamily{qhv}\selectfont \textbf{A:} purple route}} \par
    {\color{red}\footnotesize{\fontfamily{qhv}\selectfont \textbf{SAN(CNN)+STR:} 66}} \par
    {\color{red}\footnotesize{\fontfamily{qhv}\selectfont \textbf{SAAA+STR:} 508}} \par
    {\color{red}\footnotesize{\fontfamily{qhv}\selectfont \textbf{Scene Image OCR:} 1208}} \par
    {\color{orange}\footnotesize{\fontfamily{qhv}\selectfont \textbf{STR (bbox):} purple}}&

    \footnotesize{\fontfamily{qhv}\selectfont \textbf{Q:} What is the automobile sponsor of the event?}   \par {\color{blue}\footnotesize{\fontfamily{qhv}\selectfont \textbf{A:} kia}} \par
    {\color{green}\footnotesize{\fontfamily{qhv}\selectfont \textbf{SAN(CNN)+STR:} kia}} \par
    {\color{green}\footnotesize{\fontfamily{qhv}\selectfont \textbf{SAAA+STR:} kia}} \par
    {\color{orange}\footnotesize{\fontfamily{qhv}\selectfont \textbf{Scene Image OCR:} kin}} \par
    {\color{red}\footnotesize{\fontfamily{qhv}\selectfont \textbf{STR (bbox):} 0}}  &

    \footnotesize{\fontfamily{qhv}\selectfont \textbf{Q:} Which dessert is showcased?}   \par {\color{blue}\footnotesize{\fontfamily{qhv}\selectfont \textbf{A:} donut}} \par
    {\color{blue}\footnotesize{\fontfamily{qhv}\selectfont \textbf{A:} Vegan Donut}} \par
    {\color{red}\footnotesize{\fontfamily{qhv}\selectfont \textbf{SAN(CNN)+STR:} t}} \par
    {\color{orange}\footnotesize{\fontfamily{qhv}\selectfont \textbf{SAAA+STR:} Donuts}} \par
    {\color{red}\footnotesize{\fontfamily{qhv}\selectfont \textbf{Scene Image OCR:} 175}} \par
    {\color{red}\footnotesize{\fontfamily{qhv}\selectfont \textbf{STR (bbox):} north}} &

    \footnotesize{\fontfamily{qhv}\selectfont \textbf{Q:} What is preheat oven temperature?}  \par {\color{blue}\footnotesize{\fontfamily{qhv}\selectfont \textbf{A:} 350}} \par
    {\color{green}\footnotesize{\fontfamily{qhv}\selectfont \textbf{SAN(CNN)+STR:} 350 }} \par
    {\color{red}\footnotesize{\fontfamily{qhv}\selectfont \textbf{SAAA+STR:} 0}} \par
    {\color{red}\footnotesize{\fontfamily{qhv}\selectfont \textbf{Scene Image OCR:} high}} \par
    {\color{red}\footnotesize{\fontfamily{qhv}\selectfont \textbf{STR (bbox):} receivables }} 
\end{tabular}
\vspace{-0.4cm}
\caption{Qualitative results for different methods on task 1 (strongly contextualised) of the ST-VQA dataset. For each image we show the question (Q), ground-truth answer (blue), and the answers provided by different methods (green: correct answer, red: incorrect answer, orange: incorrect answer in terms of accuracy but partially correct in terms of ANLS ($0.5 \leq ANLS < 1$)).}
\label{fig:main2}
\end{center}
\end{figure*}

It is worth noting that the proposed ANLS metric generally tracks accuracy, which 
indicates broad compatibility between the metrics. 
But, in addition, 
ANLS can deal with border cases (i.e. correct intended responses, but slightly wrong recognized text) where accuracy, being a hard metric based on exact matches, cannot. Such border cases are frequent due to errors at the text recognition stage. Examples of such behaviour can be seen in the qualitative results shown in~\autoref{fig:main2} for some of the answers (indicated in orange color).
This also explains why 
the ``Scene Image OCR'' model
is better ranked in terms of ANLS than of accuracy in \autoref{tab:baseline_results}.

Finally, we notice that standard VQA models, disregarding any textual information, perform worse or comparable at best to the ``STR (retrieval)'' or ``Scene Image OCR'' models, despite the fact that these heuristic methods do not take into account the question. This observation confirms the necessity of leveraging textual information as a way to improve performance in VQA models. We demonstrate this effect by slightly improving the results of VQA models (SAAA and SAN) by using a combination of visual features and PHOC-based textual features (see SAAA+STR and SAN+STR baselines descriptions for details).


For further analysis of the baseline models' outputs and comparison between them, we provide in~\autoref{fig:results_by_type} two bar charts with specific results on different question types. In most of them the STR model is better than the ``Scene Image OCR'' (ST-OCR) in terms of ANLS. The effect of PHOC embedding is especially visible on the SAN model for correctly answering the question type such as ``what year'', ``what company'' and ``which''. Also, none of the models is capable of answering the questions regarding license plates, ``who'' and ``what number''. This is an inherent limitation of models treating VQA as a pure classification problem, as they can not deal with out of vocabulary answers. 
In this regard the importance of using PHOC features lies in their ability to capture the morphology of words rather than their semantics as in other text embeddings~\cite{mikolov2013distributed,pennington2014glove, bojanowski2017enriching}; since several text instances and answers in the dataset may not have any representation in a pre-trained semantic model. The use of a morphological embedding like PHOC can provide a starting point for datasets that contain text and answers in several languages and out of dictionary words such as license plates, prices, directions, names, etc. 


\section{Conclusions and Future Work}

This work introduces a new and relevant dimension to the VQA domain. We presented a new dataset for Visual Question Answering, the Scene Text VQA, that aims to highlight the importance of properly exploiting the high-level semantic information present in images in the form of scene text to inform the VQA process. The dataset comprises questions and answers of high variability, and poses extremely difficult challenges for current VQA methods. We thoroughly analysed the ST-VQA dataset through performing as series of experiments with baseline methods, which established the lower performance bounds, and provided important insights. Although we demonstrate that adding textual information to generic VQA models leads to improvements, we also show that ad-hoc baselines (e.g. OCR-based, which do exploit the contextual words) can outperform them, reinforcing the need of different approaches. Existing VQA models usually address the problem as a classification task, but in the case of scene text based answers the number of possible classes is intractable. Dictionaries defined over single words are also limited. Instead, a generative pipeline such as the ones used in image captioning is required to capture multiple-word answers, and out of dictionary strings such as numbers, license plates or codes. The proposed metric, namely Average Normalized Levenshtein Similarity is better suited for generative models compared to evaluating classification performance, while at the same time, it has a smooth response to the text recognition performance. 
\subsection*{Acknowledgments}
This work has been supported by projects TIN2017-89779-P,
Marie-Curie (712949 TECNIOspring PLUS), aBSINTHE (Fundacion BBVA 2017), the CERCA Programme / Generalitat de Catalunya, a European Social Fund grant (CCI: 2014ES05SFOP007), NVIDIA Corporation and PhD scholarships from AGAUR (2019-FIB01233) and the UAB.

{\small
\bibliographystyle{ieee_fullname}
\bibliography{egbib}
}

\end{document}